\pdfoutput=1

\documentclass[11pt]{article}
\usepackage{graphicx}
\usepackage{authblk}

\usepackage[]{acl}

\usepackage{times}
\usepackage{latexsym}
\usepackage{amsmath}
\usepackage{comment}

\usepackage[T1]{fontenc}

\usepackage{adjustbox}

\usepackage[utf8]{inputenc}

\usepackage{microtype}

\title{Doctor XAvIer: Explainable Diagnosis on Physician-Patient Dialogues and XAI Evaluation}

\author[1,2]{Hillary Ngai}
\author[1,2,3]{Frank Rudzicz}
\affil[1]{Department of Computer Science, University of Toronto}
\affil[2]{Vector Institute for Artificial Intelligence}
\affil[3]{Unity Health Toronto}
\affil[ ]{\texttt{hngai@cs.toronto.edu}, \texttt{frank@spoclab.com}}

\begin{document}
\maketitle
\begin{abstract}

We introduce Doctor XAvIer \textemdash a BERT-based diagnostic system that extracts relevant clinical data from transcribed patient-doctor dialogues and explains predictions using feature attribution methods. We present a novel performance plot and evaluation metric for feature attribution methods \textemdash Feature Attribution Dropping (FAD) curve and its Normalized Area Under the Curve (N-AUC). FAD curve analysis shows that integrated gradients outperforms Shapley values in explaining diagnosis classification. Doctor XAvIer outperforms the baseline with 0.97 F1-score in named entity recognition and symptom pertinence classification and 0.91 F1-score in diagnosis classification.

\end{abstract}

\section{Introduction}
Previous studies have shown that electronic medical record (EMR) data are difficult to use in machine learning systems due to the lack of regulation in data quality \textemdash EMR data are often incomplete and inconsistent \citep{amiajnl-2011-000681, doi:10.1177/1062860609336627}. Recently, there have been attempts to improve automated clinical note-taking by extracting relevant information directly from physician-patient dialogues \citep{autoscribe-jeblee, kazi-kahanda-2019-automatically, du-etal-2019-extracting}. This can alleviate physicians of tedious data entry and ensures more consistent data quality \citep{CollierE1405}.

Due to the potential in reducing costs associated with collecting patient information and diagnostic errors, there is increasing interest in using information extraction techniques in automatic diagnostic systems \citep{DBLP:journals/corr/abs-1901-10623, wei-etal-2018-task}. \citet{jeblee-etal-2019-extracting} introduced a system that extracts pertinent medical information from clinical conversations for automatic note taking and diagnosis. However, their methodology did not explore state-of-the-art natural language processing (NLP) techniques \textemdash entity extraction was done by searching the transcript for entities from medical lexicons and tf-idf was used for text classification. Although there is existing work that employs more sophisticated NLP techniques to patient-physician dialogues \citep{DBLP:journals/corr/abs-2007-07151, DBLP:journals/corr/abs-1912-04961}, there is a lack of end-to-end diagnostic systems that employ such techniques. Furthermore, all of the previous works mentioned fail to address the black-box nature of deep learning in the medical industry. Most physicians are reluctant to rely on opaque, AI-based medical technology \textemdash especially in high-risk decision-making involving patient well-being \citep{ethical-ai-healthcare}.

\begin{table}
\centering
\resizebox{\linewidth}{!}{
\begin{tabular}{ll}
    \hline
    \textbf{Speaker} & \textbf{Utterance} \\ 
    \hline
    \textbf{DR} & So how are you feeling [PATIENT NAME]? \\ 
    & \emph{O O O O O O} \\ 
    \hline
    \textbf{PT} & Not good. I'm having back and neck pain. \\
    & \emph{O O O O B-symptom O B-symptom I-symptom} \\
    \hline
    \textbf{DR} & And when did this start? \\ 
    & \emph{O B-time-expr O O B-time-expr} \\
    \hline
    \textbf{PT} & Around three days ago. \\ 
    & \emph{O B-time-expr I-time-expr I-time-expr}\\
    \hline
    \textbf{DR} & I see. Do you take any pain killers? \\ 
    & \emph{O O O O O O B-medication I-medication} \\
    \hline
    \textbf{PT} & Yes, acetaminophen and ibuprofen. \\ 
    & \emph{O B-medication O B-medication} \\
    \hline
    \end{tabular}}
\caption{Synthetic physician-patient dialogue with IOB labels. The IOB labels are italicized underneath each utterance. The \emph{B-} prefix indicates that the token is the beginning of an entity label, the \emph{I-} prefix indicates that the token is inside the entity label, and the \emph{O} indicates that the token belongs to no entity label.}
\label{tab:synthetic-dialogue}
\end{table}

In this work, we present Doctor XAvIer \textemdash a BERT-based diagnostic system that extracts relevant clinical data from transcribed patient-doctor dialogues and explains predictions using feature attribution methods. Feature attribution methods are explainable AI (XAI) methods that compute an attribution score for each input feature to represent its contribution to the model's prediction. We report feature attribution scores using integrated gradients (IG) \citep{sundararajan2017axiomatic} and Shapley values \citep{DBLP:journals/corr/LundbergL17} to provide insight as to which features are important in diagnosis classification. Descriptions of integrated gradients and Shapley values are provided in Appendix \ref{appendix:feature-attribution-methods}. Feature attribution scores could potentially help physicians build confidence in the model’s prediction or give additional insight about the relationships between different diseases and relevant patient information \citep{MARKUS2021103655}. Finally, we present a novel performance plot and evaluation metric for feature attribution methods \textemdash the Feature Attribution Dropping (FAD) curve and its Normalized Area Under the Curve (N-AUC).

\section{FAD Curve Analysis}
We introduce Feature Attribution Dropping (FAD) curve analysis for evaluating feature attribution methods. FAD curve analysis requires no modifications to the original machine learning model and is simple to implement.

\subsection{FAD Curve}
\begin{figure}
\centering
\caption{Example of an idealized FAD curve with $\beta$=20. The maximum FAD Curve AUC bounded from 0\% to $\beta$\% is shaded in pink. The actual FAD curve AUC bounded from 0\% to $\beta$\% is shaded in blue and overlaps the pink area. The N-AUC is the ratio of the blue area to the pink area.}
\label{fad-curve-auc}
\includegraphics[width=1\linewidth]{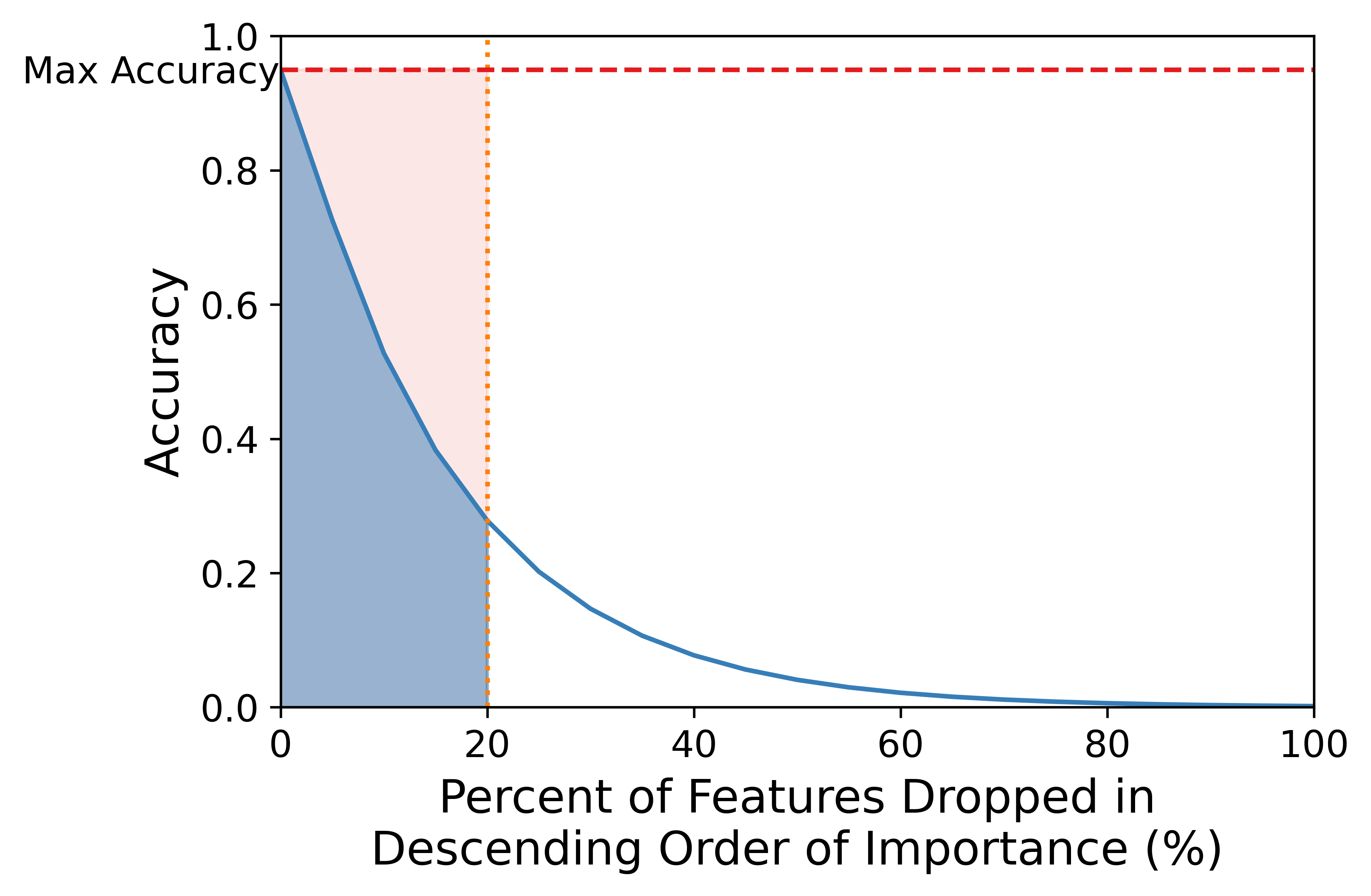}
\end{figure}

The FAD curve illustrates the explainability of a feature attribution method by plotting the performance metric (e.g., accuracy) against the percentage of features dropped in descending order of importance ranked by the feature attribution method (see Fig.~\ref{fad-curve-auc}). We define the feature importance as the absolute value of the feature attribution score to represent the magnitude of the contribution of each feature to the model's prediction. Features are dropped by modeling the absence of such features in the input. For standard machine learning inputs, continuous features can sometimes be set to their means or image pixels can sometimes be set to black \citep{sundararajan2017axiomatic}. A careful consideration of the nature of the data is, of course, required beforehand.

The intuition behind FAD curves is inspired by counterfactual explanations \textemdash which describes how the prediction of a model changes when the input is perturbed \citep{wachter2018counterfactual} \textemdash and the Pareto principle \textemdash which states that for many situations, approximately 80\% of the outcome is due to 20\% of causes (the "vital few") ~\citep{pareto1964cours, pareto-feature-attribution}. If a feature attribution method accurately ranks the most important features for a certain prediction and the Pareto principle holds true, then cumulatively dropping the most important features in descending order should yield a smaller and smaller decrease in model performance for that prediction. In other words, the model's ability to make correct predictions is mostly attributed to a small subset of important features. This entails that the steeper the FAD curve is early on, the better the feature attribution method.

\subsection{N-AUC}
We present the FAD curve Normalized Area Under the Curve (N-AUC) as a performance metric for feature attribution methods. An intuitive way to quantify how much the FAD curve decreases early on is to calculate the Area Under the Curve (AUC) bounded from 0\% to $\beta$\% of features dropped in descending order of importance. We choose $\beta$=20 using the Pareto principle, but this number is just an estimate.

Since steeper FAD curves have smaller AUCs, FAD curves with smaller AUCs indicate a better feature attribution method than FAD curves with larger AUCs. The area under the curve is approximated using the trapezoidal rule \citep{tai1994mathematical}, as described in Appendix \ref{appendix:trapezoidal-rule}. Although any performance metric can be used for FAD Curve analysis, we will use accuracy in our explanation for the sake of simplicity. The range of the FAD curve AUC is $(0, \;\beta \times max(accuracy)]$ where \begin{math} max(accuracy) \end{math} is the maximum FAD curve accuracy of all the feature attribution methods for a model's prediction and $\beta$ is the x-axis upper bound. Note that the minimum FAD curve AUC can only equal zero if the model performance is zero in the bounded range. This case is excluded from FAD curve analysis since this scenario is rare and uninformative. In order to easily compare feature attribution methods, we normalize the FAD curve AUC:

\begin{equation}
\label{n-auc}
\mathchardef\mhyphen="2D
N \mhyphen AUC = \frac{AUC}{\beta \times max(accuracy)}
\end{equation} Thus, the range of the FAD curve N-AUC is $(0, \; 1]$.

\section{Methods and Experiments}
We introduce Doctor XAvIer \textemdash a medical diagnostic system composed of joint Named Entity Recognition (NER) and intent (i.e. symptom pertinence) classification, primary diagnosis classification, and FAD curve analysis. In this section we discuss each component in detail and evaluate each component.

\subsection{Dataset}
The Verilogue dataset \citep{jeblee-etal-2019-extracting} is a collection of 800 physician-patient dialogues as audio files and their corresponding human-generated transcripts with speaker labels. Each dialogue includes the patient's information as well as the primary diagnosis. The distribution of the primary diagnoses in the dataset is shown in Appendix \ref{appendix:dataset}. The patient's information consists of the patient's age, gender, height, weight, blood pressure, smoking status, employment status, and ongoing treatments. Entities \textemdash including symptoms, medications, anatomical locations, time expressions, and therapies \textemdash are annotated by physicians in each transcript. Additional details about the dataset can be found in \citet{jeblee-etal-2019-extracting}.

\subsection{Joint NER and Intent Classification}
A diagnosis requires relevant clinical entities and a measure of pertinence of such entities. For example, a patient might mention a relevant symptom that was experienced by someone else and therefore not pertinent to diagnosis. For each sequence in the physician-patient dialogue, we extract clinical entities with NER and classify the intent of the speaker. We identify the clinical entities identified in Table \ref{ner-results}. We label each word in each sequence in the dataset using the Inside-Outside-Beginning (IOB) format ~\citep{ramshaw-marcus-1995-text}. In this paper, we focus on identifying the pertinence of symptoms. We define the intents of the patient as: \emph{confirm/deny/unsure of symptom} and the intent of both the patient and physician as: \emph{closing} (i.e., ending the conversation). Of the 407 annotated dialogues we randomly select 40 to use as a test set for NER and intent classification.

We fine-tune Bio+Clinical BERT ~\citep{DBLP:journals/corr/abs-1904-03323} jointly on these two classification tasks. This model was initialized from BioBERT ~\citep{DBLP:journals/corr/abs-1901-08746} and trained on all notes from MIMIC-III ~\citep{johnson2016mimic} \textemdash a database containing electronic health records from ICU patients. Language models pre-trained on domain-specific text yield improvements on clinical NLP tasks as compared to language models pre-trained on a general corpus ~\citep{DBLP:journals/corr/abs-2012-15419}. Since a majority of interactions between the physician and patient in the dataset are in question-and-answer format, it is beneficial to concatenate the previous sequence with the current sequence, including the respective speaker codes, to give more context to the model. This is done for each sequence before tokenization and improves NER accuracy from 89\% to 96\%.

For NER, we concatenate the last four hidden layers of Bio+Clinical BERT and feed this representation into an output layer for token-level classification. For intent classification, we feed the \texttt{[CLS]} representation of Bio+Clinical BERT into an output layer for sequence classification. We train with a batch size of 16 sequences and a maximum sequence length of 128 tokens for 5 epochs and select the model with the lowest validation loss. We use AdamW with learning rate of 2e-5, $\beta_1$ = 0.9, $\beta_2$ = 0.999, L2 weight decay of 0.01, and linear decay of the learning rate ~\citep{DBLP:journals/corr/abs-1711-05101}. We use a dropout probability of 0.1 on all layers except the output layers. 

For the loss function, we propose a linear interpolation between the intent classification Cross-Entropy (CE) loss and the average NER Negative Log Likelihood (NLL) loss with $\alpha = 0.5$. The intent classification CE loss is defined as:

\begin{equation}
\label{eq:intent-loss}
\mathcal{L}_{1}(f_{1}(\boldsymbol{x}; \boldsymbol{\theta}), \boldsymbol{y}_1) = -\sum_{i=1}^{N} y_{1, i} log f_{1, i}(\boldsymbol{x}_i;\boldsymbol{\theta})
\end{equation} where $f_{1, i}(\boldsymbol{x}; \boldsymbol{\theta})$ is the ith element of the softmax output of the intent classes, $\boldsymbol{y}_{1, i}$ is the ith element of the one-hot-encoded intent label, $N$ is the number of intent classes, $\boldsymbol{x}$ is the input, and $\boldsymbol{\theta}$ is the set of model parameters. The average NER NLL loss is defined as:

\begin{equation}
\label{eq:ner-loss}
\mathcal{L}_{2}(f_{2}(\boldsymbol{x}; \boldsymbol{\theta}), \boldsymbol{y}_2) = -\frac{\sum_{j=1}^{M} log f_{2, j}(\boldsymbol{x}_j;\boldsymbol{\theta})}{M}
\end{equation} where $f_{2, j}(\boldsymbol{x}; \boldsymbol{\theta})$ is the softmax output of the entity classes \textemdash for each token in the sequence \textemdash at the target class $j$, $\boldsymbol{y}_2$ is the set of entity labels, and $M$ is the number tokens in the sequence. The full loss function is defined in Appendix \ref{appendix:loss-function-equations}. \texttt{[PAD]} tokens are excluded from the loss function using masking.

As seen in Table \ref{ner-results} and Table \ref{intent-results}, the model yields approximately 0.97 weighted precision, recall, and F1-score on both tasks, outperforming \citet{jeblee-etal-2019-extracting}'s models. However, the exact results are difficult to compare since \citet{jeblee-etal-2019-extracting} tested their model on a smaller subset of the dataset.

\begin{table}
\centering
\resizebox{\linewidth}{!}{
\begin{tabular}{lllll}
    \hline
    \bfseries Entity & \bfseries Instances & \bfseries P & \bfseries R & \bfseries F1 \\
    \hline
    Other & 158,018 & 0.98 & 0.98 & 0.98\\
    Anatomical Location & 598 & 0.73 & 0.65 & 0.69\\
    Bodily Function & 6 & 0.00 & 0.00 & 0.00\\
    Diagnosis & 1,345 & 0.79 & 0.75 & 0.77\\
    Therapy & 1420 & 0.62 & 0.69 & 0.65\\
    Medication & 3,324 & 0.90 & 0.81 & 0.85\\
    Referral & 256 & 0.71 & 0.79 & 0.74\\
    Symptom & 3,574 & 0.57 & 0.66 & 0.61\\
    Substance Use & 68 & 0.00 & 0.00 & 0.00\\
    Time Expression & 4,062 & 0.90 & 0.84 & 0.87\\
    \hline
    Weighted Avg & 172,671 & 0.97 & 0.96 & 0.97 \\
    \hline
    \end{tabular}}
\caption{Named entity recognition results.}
\label{ner-results}
\end{table}

\begin{table}
\centering
\resizebox{\linewidth}{!}{
\begin{tabular}{lllll}
    \hline
    \bfseries Intent & \bfseries Instances & \bfseries P & \bfseries R & \bfseries F1 \\
    \hline
    Confirm Symptom & 228 & 0.70 & 0.69 & 0.70\\
    Deny Symptom & 52 & 0.73 & 0.69 & 0.71\\
    Unsure of Symptom & 73 & 0.34 & 0.65 & 0.62\\
    Closing & 28 & 0.29 & 0.47 & 0.36\\
    Other & 6,425 & 0.99 & 0.99 & 0.99\\
    \hline
    Weighted Avg & 6,806 & 0.97 & 0.97 & 0.97 \\
    \hline
    \end{tabular}}
\caption{Intent classification results.}
\label{intent-results}
\end{table}

\subsection{Primary Diagnosis Classification}

We classify the primary diagnosis for each physician-patient dialogue using the the patient's information \textemdash such as the patient's age, weight, blood pressure, and smoking status \textemdash and the extracted symptoms from the conversation. Since the same symptom can be said in various different ways, we compile a set of symptoms of all the diseases in the dataset according to WedMD and assign each extracted symptom to one of the pre-defined symptoms. We use a pre-trained SentenceBERT (SBERT) model \citep{DBLP:journals/corr/abs-1908-10084} to embed each extracted symptom and all the pre-defined symptoms. Each extracted symptom is assigned to its most similar pre-defined symptom measured by the cosine similarity between the SBERT embeddings \citep{DBLP:journals/corr/abs-2101-11432}. The most similar pre-defined symptom is defined as:
\begin{equation}
\label{eq:symptom-assignment}
s^\ast_i = \underset{s_i}{\arg\max} \; \mathrm{sim}(\mathrm{emb}(e_j), \mathrm{emb}(s_i)) \; \forall s_i \in S
\end{equation} where $S = \{s_1, ..., s_N\}$ is the set of symptoms of all diseases in the dataset, $s_i$ is the i\textsuperscript{th} symptom in $S$, $e_j$ is the j\textsuperscript{th} extracted symptom, $\mathrm{emb}(x)$ is the SBERT embedding of text $x$, and $\mathrm{sim}(a, b)$ is the cosine similarity between embeddings $a$ and $b$. The assigned pre-defined symptom is:
\begin{equation}
\label{eq:symptom-assignment-2}
\resizebox{0.9\linewidth}{!}{%
$e^\ast_j =
    \begin{cases} 
        s^\ast_i \, , \; if \; \mathrm{sim}(\mathrm{emb}(e_j), \mathrm{emb}(s^\ast_i)) \geq \epsilon \\
        None \\
    \end{cases}$%
}\\
\end{equation} where $\epsilon$ is a constant and $None$ represents that we do not use the extracted symptom $e_j$ for classification. We chose $\epsilon = 0.35$ since it minimized incorrect assignments of extracted symptoms in the dataset while filtering out less than 10\% of extracted symptoms.

The diagnosis classification model is a neural network composed of 549 input features and three hidden layers with 182K total parameters. The input features consists of patient information and the pertinence of extracted symptoms from the conversation. The model is evaluated using stratified $5$-fold cross-validation. We train with a batch size of 32 for 100 epochs and select the model with the lowest validation loss. We use Adam \citep{kingma2017adam} with learning rate of 1e-3, $\beta_1$ = 0.9, $\beta_2$ = 0.999, and $\epsilon$ = 1e-08. We use a GELU activation \citep{DBLP:journals/corr/HendrycksG16} on all hidden layers. The training loss is the standard CE loss.

As seen in Table \ref{diagnosis-results}, Doctor XAvIer yields a significant improvement in weighted precision, recall, and F1-score for diagnosis classification compared to the baseline \cite{jeblee-etal-2019-extracting}.

\begin{table}
\centering
\resizebox{\linewidth}{!}{
\begin{tabular}{lllll}
    \hline
    \bfseries Diagnosis & \bfseries Model & \bfseries P & \bfseries R & \bfseries F1 \\
    \hline
    ADHD & Doctor XAvIer & \textbf{0.95} & \textbf{0.97} & \textbf{0.96}\\
     & \citep{jeblee-etal-2019-extracting} & 0.84 & 0.84 & 0.83\\
    \hline 
    Depression & Doctor XAvIer & \textbf{0.92} & \textbf{0.93} & \textbf{0.92}\\
     & \citep{jeblee-etal-2019-extracting} & 0.80 & 0.64 & 0.71\\
    \hline 
    Osteoporosis & Doctor XAvIer & \textbf{0.85} & 0.69 & 0.75\\
     & \citep{jeblee-etal-2019-extracting} & 0.81 & \textbf{0.78} & \textbf{0.78}\\
    \hline 
    Influenza & Doctor XAvIer & \textbf{1.00} & \textbf{0.99} & \textbf{0.99}\\
     & \citep{jeblee-etal-2019-extracting} & 0.91 & 0.95 & 0.93\\
    \hline 
    COPD & Doctor XAvIer & \textbf{0.93} & \textbf{0.93} & \textbf{0.93}\\
     & \citep{jeblee-etal-2019-extracting} & 0.75 & 0.65 & 0.68\\
    \hline 
    Type II Diabetes & Doctor XAvIer & 0.52 & 0.47 & 0.48\\
     & \citep{jeblee-etal-2019-extracting} & \textbf{0.81} & \textbf{0.75} & \textbf{0.76}\\
    \hline 
    Other & Doctor XAvIer & \textbf{0.73} & 0.80 & \textbf{0.76} \\
     & \citep{jeblee-etal-2019-extracting} & 0.71 & \textbf{0.82} & \textbf{0.76}\\
    \hline 
    Weighted Avg & Doctor XAvIer & \textbf{0.91} & \textbf{0.91} & \textbf{0.91} \\
     & \citep{jeblee-etal-2019-extracting} & 0.82 & 0.80 & 0.80 \\
    \hline
    \end{tabular}}
\caption{K-fold cross-validation primary diagnosis classification results.}
\label{diagnosis-results}
\end{table}

\subsection{Evaluation of Explainability Methods}
For each test fold and model trained on the train fold in the stratified $5$-fold cross-validation of the diagnosis classification model, we evaluate each feature attribution method using FAD curve analysis. We choose accuracy as the performance metric for FAD curve analysis.

\begin{figure}
\centering
\caption{K-fold cross-validation \emph{ADHD} and \emph{Depression} FAD curves.}
\label{adhd-depression-fad-curve}
\includegraphics[width=\linewidth]{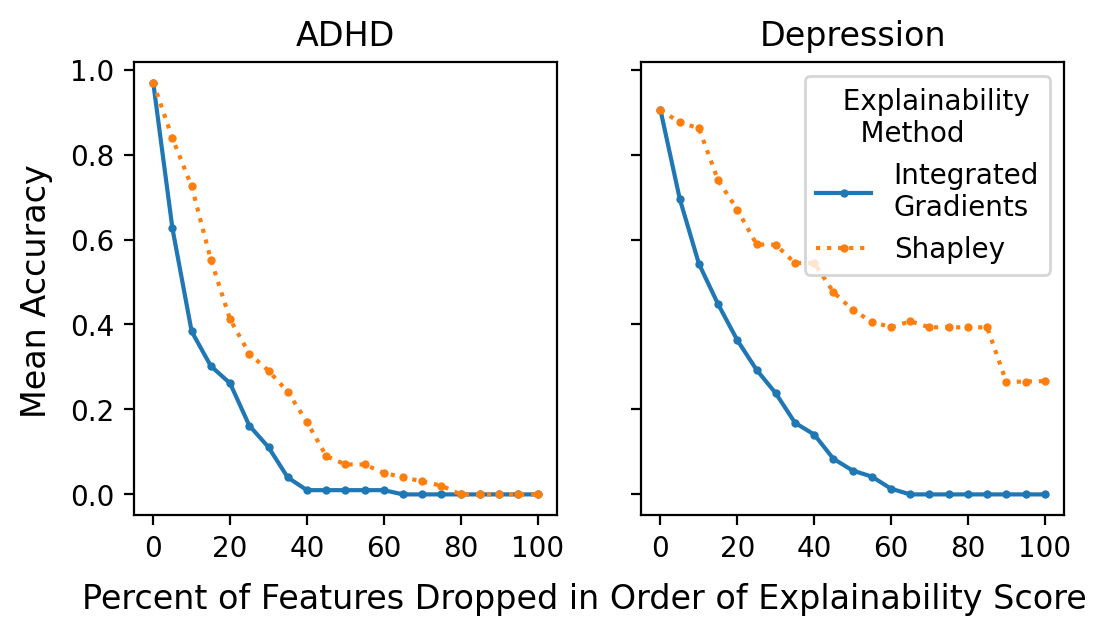}
\end{figure}

\begin{figure}
\centering
\caption{K-fold cross-validation \emph{COPD} and \emph{Type II Diabetes} FAD curves.}
\label{copd-type-ii-diabetes-fad-curve}
\includegraphics[width=\linewidth]{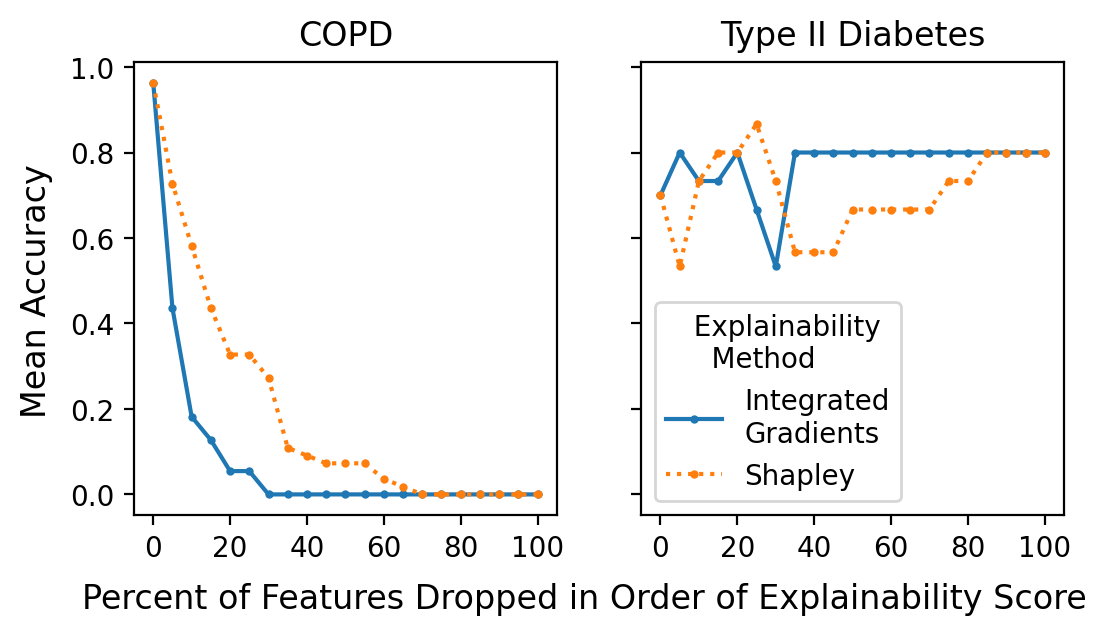}
\end{figure}

\begin{figure}
\centering
\caption{K-fold cross-validation \emph{Osteoporosis} and \emph{Influenza} FAD curves.}
\label{osteoporosis-influenza-fad-curve}
\includegraphics[width=\linewidth]{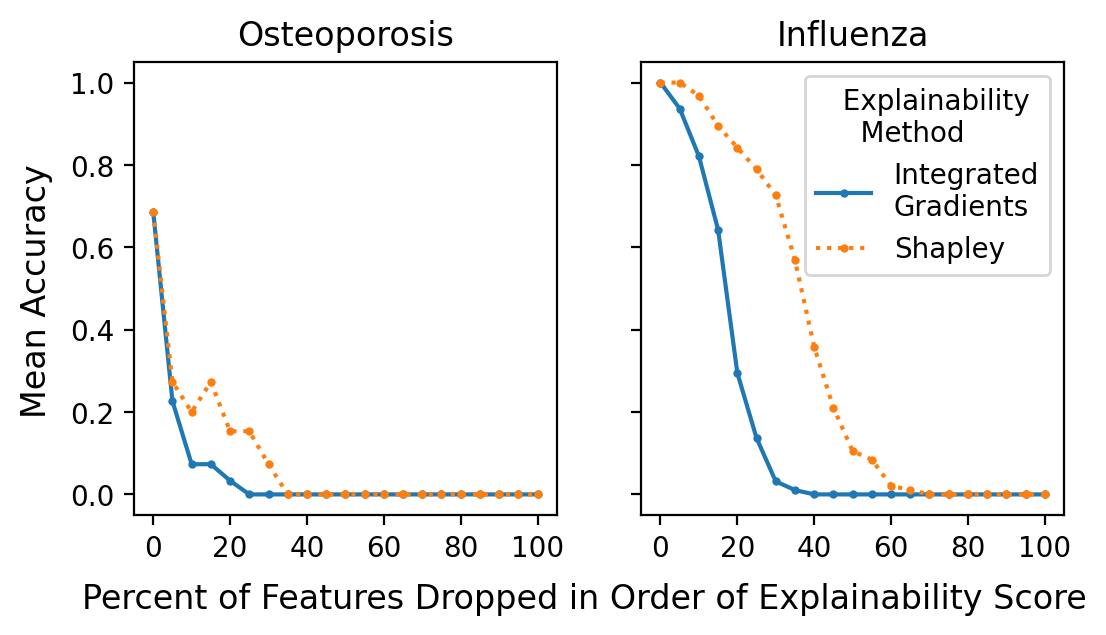}
\end{figure}

As seen in Table \ref{fad-auc-results}, integrated gradients outperforms Shapley values according to FAD curve analysis \textemdash achieving smaller N-AUCs for all diagnoses. As seen in Figures \ref{adhd-depression-fad-curve}, \ref{copd-type-ii-diabetes-fad-curve}, and \ref{osteoporosis-influenza-fad-curve} and Appendix \ref{appendix:fad-curves}, integrated gradients yields noticeably steeper FAD curves than Shapley values for all of the diagnoses except \emph{Type II Diabetes}. The sporadic shapes of the \emph{Type II Diabetes} FAD curves can potentially be explained by the lack of dialogues with \emph{Type II Diabetes} as their primary diagnosis \textemdash there are only 3 instances. This suggests that we could potentially improve performance by collecting more instances of the infrequent classes or performing regularization. 

It is important to note that some features in the dataset may be correlated. Therefore, dropping features that are correlated with other features may lead to an increase \textemdash instead of a decrease \textemdash in the performance metric despite dropping features in descending order of importance. We could potentially mitigate this by using feature selection methods before performing FAD curve analysis.

\begin{table}
\centering
\resizebox{0.85\linewidth}{!}{
\begin{tabular}{llll}
    \hline
    \bfseries Diagnosis & \bfseries Instances & \bfseries IG & \bfseries Shapley \\
    \hline
    ADHD & 20 & \textbf{0.48} & 0.77 \\
    Depression & 14 & \textbf{0.63} & 0.85 \\
    Osteoporosis & 5 & \textbf{0.24} & 0.36 \\
    Influenza & 19 & \textbf{0.72} & 0.95 \\
    COPD & 11 & \textbf{0.33} & 0.59 \\
    Type II Diabetes & 3 & \textbf{0.59} & 0.73 \\
    Other & 9 & \textbf{0.71} & 0.95 \\
    \hline
    \end{tabular}}
\caption{K-fold cross-validation FAD curve N-AUC from 0\% to 20\% of dropped features comparing integrated gradients and Shapley values.}
\label{fad-auc-results}
\end{table}

\section{Conclusion}
Doctor XAvIer yields significant improvements in NER, symptom pertinence classification, and diagnosis classification compared to previous work \citep{jeblee-etal-2019-extracting}, while also explaining why the model made each diagnosis. We also present a novel performance plot and evaluation metric for feature attribution methods \textemdash FAD curve analysis and its N-AUC. FAD curve analysis shows that integrated gradients outperforms Shapley values in explaining diagnosis classification in the Verilogue dataset. In our future work, we will calculate $\beta$ in a data-driven manner to standardize FAD curve analysis for a given dataset. We will also apply FAD curve analysis to other feature attribution methods, AI domains, and datasets to evaluate its generalizability.

\bibliography{custom}
\bibliographystyle{acl_natbib}

\section*{Appendix}
\appendix

\section{Feature Attribution Methods}
\label{appendix:feature-attribution-methods}

\subsection{Shapley Values}
The Shapley value \citep{DBLP:journals/corr/LundbergL17} \textemdash a method from cooperative game theory \textemdash assigns payouts to players depending on their contribution to the total payout in a cooperative game. Players cooperate in a coalition and receive a certain profit from this cooperation. In explainable AI, the game is the prediction task for a single instance in the dataset, the players are the feature values of a single instance that collaborate to make a prediction, and the gain is the prediction for an instance minus the average prediction for all instances \citep{DBLP:journals/corr/abs-1908-08474}. In other words, the Shapley value measures the contribution of each input feature to a model's prediction for a single instance.

\subsection{Integrated Gradients}
Integrated gradients \citep{sundararajan2017axiomatic} is an XAI technique which attributes the prediction of a deep neural network to its input features. Integrated gradients attributes blame to an input feature by using the absence of the input feature as a baseline for comparing outcomes. For most deep networks, there exists a baseline in the input space where the prediction is neutral. For example, the baseline for an object recognition network can be a black image. Mathematically, integrated gradients is defined as the path integral of the gradients along the straightline path from the baseline $x'$ to the input $x$.

\section{Area Under the Curve Approximation}
\label{appendix:trapezoidal-rule}
The area under the curve is approximated using the trapezoidal rule \citep{tai1994mathematical}:
\begin{equation}
\label{trapezoidal-rule}
\begin{split}
AUC & = \int_0^{20} f(x) \, dx \\
 & \approx \sum_{k=1}^{N} \frac{f(x_{k-1})+f(x_{k})}{2} \, \Delta x_k
\end{split}
\end{equation} where
\begin{math}
0 = x_0 < x_1 < ... < x_{N-1} < x_N = 20
\end{math}
and 
\begin{math}
\Delta x_k = x_k - x_{k-1}
\end{math}.

\section{Additional Dataset Details}
Table \ref{tab:diagnosis-distribution} shows the distribution of diagnoses in the Verilogue dataset.

\label{appendix:dataset}
\begin{table}
\centering
\resizebox{0.65\linewidth}{!}{
    \begin{tabular}{ll}
        \hline
        \textbf{Primary Diagnosis} & \textbf{Dialogues} \\ 
        \hline
        ADHD &  99\\ 
        Depression & 72 \\
        Osteoporosis & 26 \\ 
        Influenza & 95 \\ 
        COPD & 55 \\ 
        Type II Diabetes & 14 \\ 
        Other & 46 \\ 
        \hline
        \end{tabular}
        }
\caption{Distribution of primary diagnoses in the Verilogue dataset.}
\label{tab:diagnosis-distribution}
\end{table}

\section{Additional Details for Joint NER and Intent Classification}
\subsection{Loss Function Equations}
\label{appendix:loss-function-equations}

Combining Eq. \ref{eq:intent-loss} and Eq. \ref{eq:ner-loss}, the joint intent classification and NER loss function is defined as:
\begin{equation}
\label{eq:loss}
\begin{split}
& \mathcal{L}(f_{1}(\boldsymbol{x}; \boldsymbol{\theta}), \boldsymbol{y}_1, f_{2}(\boldsymbol{x}; \boldsymbol{\theta}), \boldsymbol{y}_2, \alpha) \\
& = \alpha \mathcal{L}_{1}(f_{1}(\boldsymbol{x}; \boldsymbol{\theta}), \boldsymbol{y}_1) \\ 
& + (1-\alpha) \mathcal{L}_{2}(f_{2}(\boldsymbol{x}; \boldsymbol{\theta}), \boldsymbol{y}_2)
\end{split}
\end{equation} where $\alpha \in [0, 1]$.

\subsection{Training Hardware}
Training of the joint NER intent classiciation model was performed on a NVIDIA Quadro RTX 6000 GPU and took approximately two hours to finish training.

\begin{table}
\centering
\resizebox{\linewidth}{!}{
\begin{tabular}{ll}
    \hline
    \bfseries Feature & \bfseries Attribution \% \\
    \hline
    Age & 0.015 \\ 
    Trouble making decisions and remembering things & 0.013 \\
    Taking Adderall & 0.009 \\ 
    Trouble focusing on a task & 0.007 \\
    Easily distracted & 0.004 \\
    Restlessness & 0.003 \\
    \hline
    \end{tabular}}
\caption{Examples of top features for classifying ADHD ranked by integrated gradients.}
\label{ig-adhd-examples}
\end{table}

\begin{table}
\centering
\resizebox{\linewidth}{!}{
\begin{tabular}{ll}
    \hline
    \bfseries Feature & \bfseries Attribution \% \\
    \hline
    Weight & 0.003 \\ 
    Age & 0.002 \\ 
    Trouble focusing on a task & 0.002 \\
    Trouble making decisions and remembering things & 0.002 \\ 
    Easily distracted & 0.002 \\
    Systolic Blood Pressure & 0.002 \\
    \hline
    \end{tabular}}
\caption{Examples of top features for classifying ADHD ranked by Shapley values.}
\label{shapley-adhd-examples}
\end{table}

\section{Additional Details for Primary Diagnosis Classification}
\subsection{Training Hardware}
Training of the primary diagnosis classification model was performed on a NVIDIA Tesla K80 GPU and took approximately an hour to finish training and evaluating all five models.

\section{Additional Details for FAD Curve Analysis}
\subsection{Feature Attribution Examples}
Examples of top features for classifying ADHD ranked by integrated gradients are shown in Table \ref{ig-adhd-examples} and examples of top features for classifying ADHD ranked by Shapley values are shown in Table \ref{shapley-adhd-examples}.

\subsection{Additional FAD Curves for Diagnosis Classification}
\label{appendix:fad-curves}

\begin{figure}
\centering
\caption{K-fold cross-validation \emph{Other} FAD curves.}
\label{other-fad-curve}
\includegraphics[width=.5\linewidth]{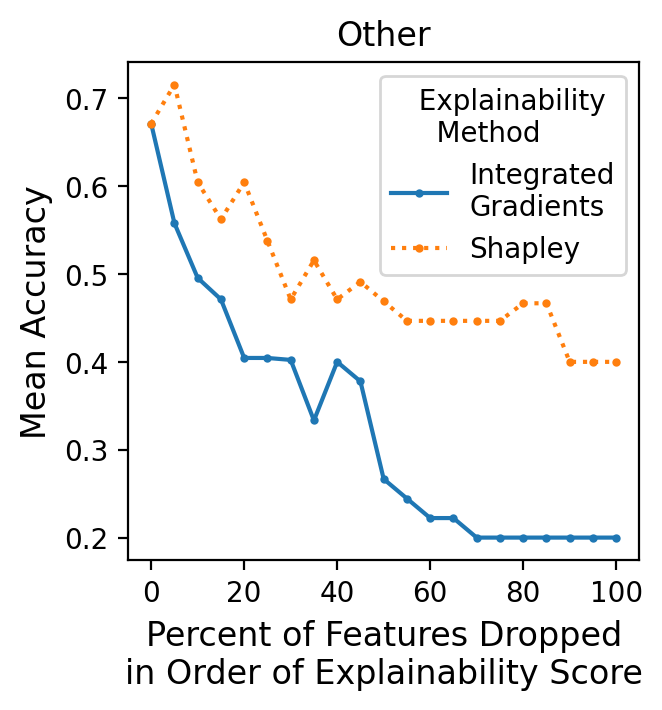}
\end{figure}

The FAD curve for the diagnosis \emph{Other} is seen in Figure \ref{other-fad-curve}.

\section{Code}
The code is available at: \url{https://github.com/hillary-ngai/doctor_XAvIer}.

\end{document}